\documentclass{article}

 \usepackage[final,nonatbib]{nips_2017}

\usepackage[utf8]{inputenc} % allow utf-8 input
\usepackage[T1]{fontenc}    % use 8-bit T1 fonts
\usepackage{hyperref}       % hyperlinks
\usepackage{url}            % simple URL typesetting
\usepackage{booktabs}       % professional-quality tables
\usepackage{amsfonts}       % blackboard math symbols
\usepackage{nicefrac}       % compact symbols for 1/2, etc.
\usepackage{microtype}      % microtypography
\usepackage{graphicx}
\usepackage{multirow}
\usepackage{subfigure}
\usepackage{url}

\title{ A deep learning-based method for relative location prediction in CT scan images}

\author{
  Jiajia Guo, Hongwei Du \thanks{Corresponding author. E-mail address:duhw@ustc.edu.cn.} , Bensheng Qiu, Xiao Liang\\
  Center for Biomedical imaging\\
  University of Science and Technology of China\\
  Hefei, Anhui 230026, China \\
}

\begin{document}
\maketitle
\begin{abstract}
Relative location prediction in computed tomography (CT) scan images is a challenging problem. 
In this paper, a regression model based on one-dimensional convolutional neural networks is proposed to determine the relative location of a CT scan image both robustly and precisely. 
A public dataset is employed to validate the performance of the study’s proposed method using a 5-fold cross validation. 
Experimental results demonstrate an excellent performance of the proposed model when compared with the state-of-the-art techniques, achieving a median absolute error of 1.04 cm and mean absolute error of 1.69 cm. 
\end{abstract}

\section{Introduction}
In the clinical diagnosis process, to assess the development of the disease, medical practitioners often need to find the differences between CT scan images from the same body region of a patient at different time or similar body regions of different patients. 
Hence, the medical picture archiving and communication system (PACS)  \cite{cooke2003picture} usually has to load the complete CT dataset for medical practitioners’ selection, which not only spends too much time but also occupies too much memory space. 
In this scenario, only several CT slice images are really needed to be loaded rather than the complete dataset. 
Thus, if the location of a CT scan image has been known, medical practitioners will directly download the needful images from the dataset. 
For the above reason, determining the location of the given axial CT scan images precisely is quite necessary in the clinical diagnosis process. 
Due to different heights of patients, relative location should be paid attention to rather than absolute location.

In the previous literature, the k-nearest neighbors (KNN) algorithm is the most commonly used algorithm \cite{emrich2010ct,graf20112d,graf2011position,fernandez2014diffusion,raytchev2016ensemble}. KNN determines the most suitable location by finding the nearest neighbors in the training set to the target CT scan image in terms of different distances (e.g. Euclidean distance and diffusion-based weighted distance). To the best of our knowledge, the method proposed in \cite{raytchev2016ensemble} outperforms other previous methods, which achieved a median error of 1.22 cm. That literature used a linear regression with a random subspace-based regularization on the k-nearest neighbors of the training set. Some other machine learning algorithms have also been applied to tackle this problem such as Genetic Programming-based algorithm \cite{castelli2016prediction} and Random Forest-based algorithm \cite{liu2013rule}.

In the past several years, the high development of convolutional neural networks (CNN), which uses multiple processing layers to learn high-level features in training data, makes it possible to learn parameters of regression networks from end to end. 
However, there are still no studies on how to build a CNN model for the relative location prediction (RLP) in CT scan images \cite{litjens2017survey}  while deep learning has achieved great success in many other domains \cite{lecun2015deep}.
In this work, we proposed a one-dimensional (1D) CNN-based regression model for the RLP in CT scan images which meets the expectations of accuracy, robustness and speed. 

\section{Materials and method}
\subsection{Dataset information}
\begin{figure}[h]
	\centering 
    \subfigure[Rescaled CT image.]{ \label{fig1:a} 
     \includegraphics[scale=0.35]{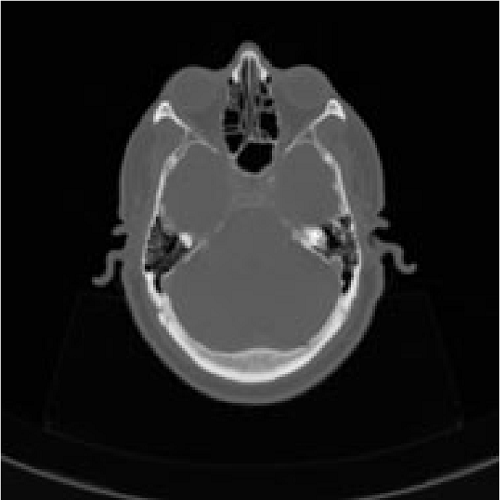} 
     } 
	\subfigure[Detected region.]{ \label{fig1:b} 
	\includegraphics[scale=0.35]{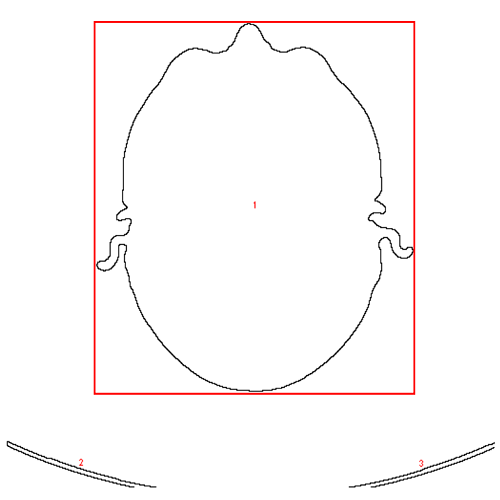} 
	}
	\subfigure[Sector/shell model.]{ \label{fig1:c} 
	\includegraphics[scale=0.35]{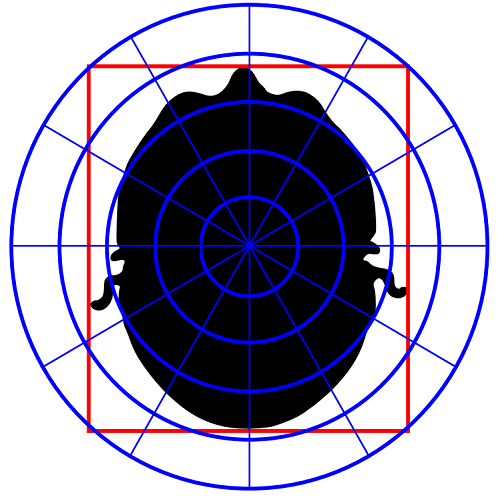} 
	}
	\caption{\label{feature extraction} The feature extraction process for a neck scan image \cite{graf20112d}.} 
\end{figure}

The dataset used in this work was retrieved from a set of 53,500 CT scan images taken from 74 different patients in the UCI repository \cite{Lichman:2013}. The feature vectors in this repository were extracted from the preprocessed images by three steps, as shown in Fig.~\ref{feature extraction}.   
First, the CT scan image (Fig.~\ref{fig1:a}) was scaled to a common resolution (1.5 px/mm) to reduce the scale variance between different CT scan images. Then, a compound region detection (Fig.~\ref{fig1:b}) was applied using a threshold in Hounsfield Unit (HU) value to separate the body region from the whole image. 
Finally, shape context algorithm \cite{belongie2001shape} extracted two specific radial descriptors, as shown in Fig.~\ref{fig1:c}. The first descriptor described the spatial distribution of bones, while the second described the location and arrangement of soft tissue inside the body. Both descriptors were combined into a 1D feature vector with 384 attributes. More details can be found in the literature \cite{graf20112d}. 

To accelerate the convergence of the weights in the network, we first normalized the feature vectors via z-score normalization. After this normalization, all sample vectors have a mean of 0 and a standard deviation of 1. 

\subsection{1D convolutional neural network regression model}
Recently, many researchers have successfully applied CNN to 1D data. For example, to predict the functional output of DNA sequences, the literature \cite{kelley2016basset} designed a deep 1D-CNN model whose performance outperformed all traditional methods. This result greatly encourages us to introduce 1D-CNN into tackling the RLP problem in CT scan images. In the following subsections, we show the network architecture and the loss function used in this work.

\subsubsection{Network architecture}
\begin{figure}[h]
  \centering
  \includegraphics[scale=0.5]{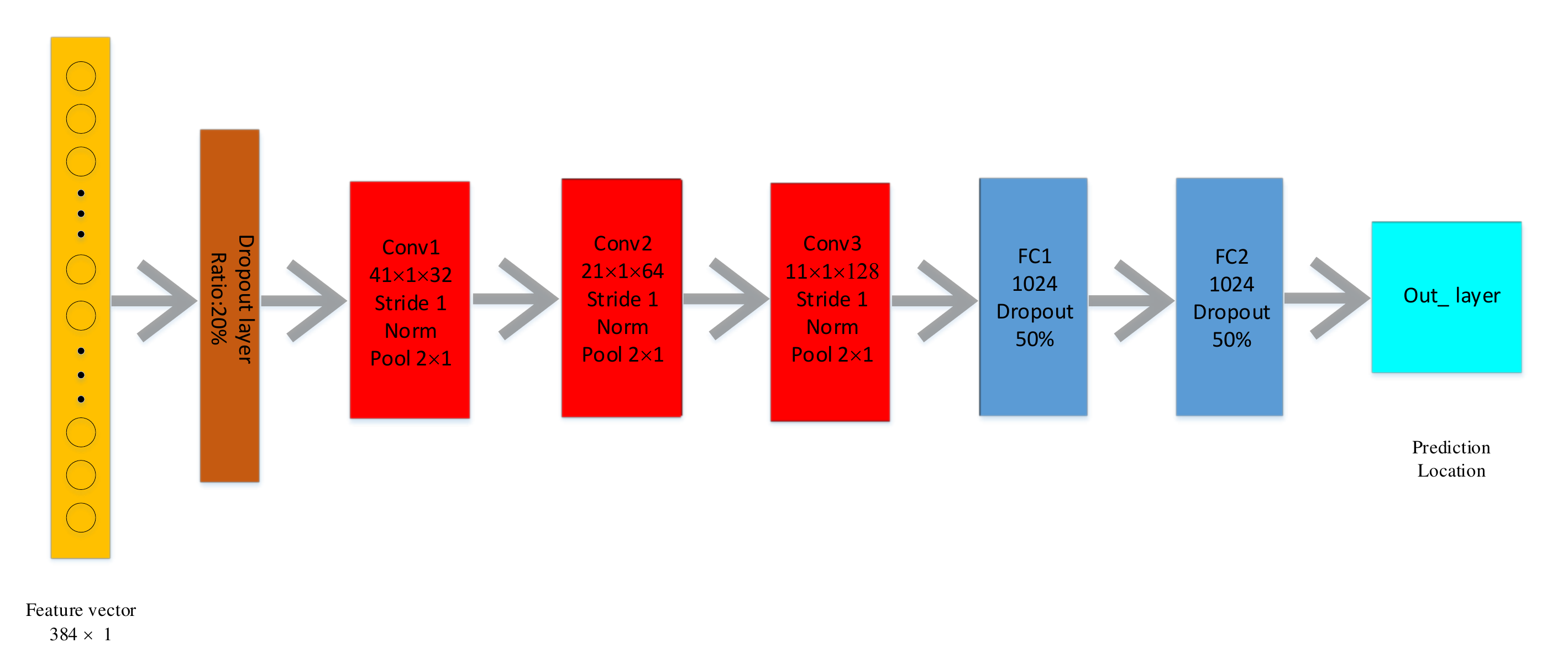}
  \caption{\label{architecture} Overview of a 1D-CNN regression model architecture.}
\end{figure}

The detailed configuration of this 1D-CNN model is shown in Fig. ~\ref{architecture}. The input of the network is a 1D feature vector extracted by shape context algorithm. The first layer is a dropout layer with 20\% ratio of dropped outputs. Then, the first convolutional layer (Conv1) has 32 $41\times1$ filters with a stride 1, followed by a batch normalization (BN) layer \cite{ioffe2015batch} and a $2\times1$ max pooling layer with a stride 2. The second convolutional layer (Conv2) has 64 $21\times1$ filters with a stride 1, followed by a BN layer and a $2\times1$ max pooling layer with a stride 2. The last convolutional layer (Conv3) has 128 $11\times1$ filters with a stride 1, followed by a BN layer and a $2\times1$ max pooling layer with a stride 2. Then, the following two layers are two 1024–D fully-connected layers (FC1 and FC2), both of which are followed by a dropout layer with 50\% ratio of dropped outputs. As a regression network, the last layer is a 1-D fully-connected layer which outputs the estimated relative location for the input feature vector. All activation functions of these layers are the Rectified Linear Units (ReLU) function \cite{krizhevsky2012imagenet}. The number of parameters in this network is about 5.5M.

\subsubsection{Loss function}
\label{loss}
Loss function is one of the most important factors in deep learning, since the goal of an optimization problem can be regarded as a process of minimizing the proposed loss function. 
When the task is a regression problem, the Mean Squared Error (MSE) loss is the nearly ubiquitous preference of deep learning researchers for its good mathematical properties such as convexity and being strictly differentiable.
And to prevent overfitting in the training process, L2 regularization was added to the loss function.
\section{Experiments and results}
\subsection{Implementation details}
In the experiments, to ensure all samples in the dataset to be used for both training and testing, we applied a 5-fold cross validation where the folds were split by patient. The proposed model was implemented in Tensorflow and trained by the Adam optimizer \cite{kingma2014adam} with the MSE loss. The remaining hyper parameters of the network are shown in Table~\ref{Hyper parameters}. The experiments were run on an image processing workstation with Intel Xeon E5-2630 @ v4 2.20GHz and GPU ASUS GeForce @ GTX 1080 Ti Founders Edition 11GB GDDR5X.

\begin{table}[th]
  \caption{Hyper parameters of the proposed method.}
  \label{Hyper parameters}
  \centering
  \begin{tabular}{lll}
    \toprule
    Stage     & Hyper-parameter     & Value \\
    \midrule    
    \multirow{2}*{Initialization of the convolutional layer} & Bias  & 0  \\
          & Weight	& Gaussian distribution    \\
    \multirow{3}*{Learning rate}&	Base learning rate&	0.0001 \\
              &Decay step	 &20000\\
            &Decay rate       & 0.5 \\
    \multirow{3}*{Adam optimizer} &	Beta1	& 0.9\\
	          &Beta2	& 0.999\\
	        &Epsilon	&  $10^{-8}$  \\
    Training&	Batch size	&256      \\
    \bottomrule
  \end{tabular}
\end{table}

\subsection{ Evaluation metrics}
The most widely used evaluation metric for RLP problem is the median absolute error (MdAE) for its high robustness to outliers. 
This evaluation metric has been employed in the previous works  \cite{fernandez2014diffusion,raytchev2016ensemble,castelli2016prediction}, and in order to compare with previous methods directly, we use this metric as a main metric in this paper. 
In the \cite{graf20112d,liu2013rule}, the mean absolute error (MAE) and relative root mean square error (RRMSE) were also used as the evaluation metric. So, to compare with them, MAE and RRMSE are also used as additional metrics.
\subsection{Experimental results}

\subsubsection{Comparison with traditional artificial neural networks}
In the literature \cite{castelli2016prediction}, a method based on traditional artificial neural network (ANN) algorithm \cite{haykin1994neural} was applied to the RLP problem. The MdAE of the ANN-based method was up to 15.32cm,while the MdAE of 1D-CNN model is just 1.04 cm. The reason behind this performance may be as follows: the feature vectors used in this work were extracted by shape context algorithm \cite{belongie2001shape}, and the nearby elements in the extracted feature vectors had the spatial correlations. 
However, traditional ANN ignores these correlations and just regards the elements of the input vectors as independent elements. 
To the opposite, the CNN-based method takes advantage of the signal correlation by using local connections, shared weights and pooling \cite{lecun2015deep}. Thus, CNN can perform much better in the RLP problem than traditional ANN.
\subsubsection{Comparison with other traditional machine leaning techniques}
To the best of our knowledge, the literatures \cite{graf20112d,fernandez2014diffusion,raytchev2016ensemble,castelli2016prediction,liu2013rule} used the same dataset as this research work. Thus, we compare the result of 1D CNN-based method with that of them. Table~\ref{Experimental comparison} shows that the proposed method performs better than any other method in any evaluation metric. 
\begin{table}[h]
  \caption{Experimental comparison for the accuracies using different techniques. ‘-’ represents the results are not reported in these papers.}
  \label{Experimental comparison}
  \centering
  \begin{tabular}{lclclclclcl}
    \toprule
    Method   &Time  & MdAE     & MAE &RRMSE & Improvement \\
    \midrule 
    KNN\cite{graf20112d}&2011 & -	& 1.80	& -	& 6.1\%  \\
	Random Forest\cite{liu2013rule}&2013 & -	& -	& 0.28	& 46.4\%  \\
	Anisotropic diffusion KNN \cite{fernandez2014diffusion}&2014& 1.65	& -	& -	& 37.0\%  \\
	Random subspace KNN \cite{raytchev2016ensemble}&2016 & 1.22	& -	&-	& 14.8\%  \\
	Local search Genetic programming \cite{castelli2016prediction}&2016 & 3.44& -	& -	& 69.8\%  \\
	\textbf{Ours}&\textbf{2017} &\textbf{1.04} &\textbf{1.69} & \textbf{0.15} & \textbf{-} \\
    \bottomrule
    \end{tabular}
\end{table}

As can be seen in Table ~\ref{Experimental comparison}, the most widely used method is based on KNN. KNN algorithm is a lazy learning method with low efficiency and its performance heavily depends on the selection of a “good value” for the number of the nearest neighbors\cite{saeys2007review}. Due to these two shortcomings, it is difficult to apply the KNN-based method to medical applications. To the opposite, the proposed method can overcome these shortcomings. The parameters of this model are learned from end to end. And it takes just about 3.5 ms for each online RLP though the offline training of the model needs about 15 min. 

\section{Discussion and conclusion}
In this paper, a 1D-CNN regression model for the RLP in CT scan images was presented.We experimentally demonstrate that our proposed method achieves superior performance compared with other several methods using the same dataset and explain the reason behind the excellent performance of 1D-CNN. To the best of our knowledge, it is also the first time that deep learning has been applied to this research. 

In the future, the proposed method will be applied to a bigger dataset which consists of enough CT scans, and extend the scan to the whole human body. 
\subsubsection*{Acknowledgments}
This work is supported by Information Science Center, University of Science and Technology of China and the National Key Scientific Instrument and Equipment Development Projects of China [81527802].
{
\small
\bibliographystyle{unsrt}
\bibliography{reffile}
}
\end{document}